\documentclass[final]{cvpr}

\usepackage{times}
\usepackage{epsfig}
\usepackage{graphicx}
\usepackage{amsmath}
\usepackage{amssymb}
\usepackage{float}

\usepackage[pagebackref=true,breaklinks=true,colorlinks,bookmarks=false]{hyperref}

\def\cvprPaperID{****} % *** Enter the CVPR Paper ID here
\def\confYear{CVPR 2021}

\begin{document}

%%%%%%%%% TITLE
\title{Trajectory Prediction using Generative Adversarial Network in Multi-Class Scenarios}

\author{Shilun Li\\
{\tt\small shilun@stanford.edu}
\and
Tracy Cai\\
{\tt\small cpcai@stanford.edu}
\and
Jiayi Li\\
{\tt\small jiayili@stanford.edu}
}

\maketitle

%%%%%%%%% ABSTRACT
\begin{abstract}

Predicting traffic agents’ trajectories is an important task for auto-piloting. Most previous work on trajectory prediction only considers a single class of road agents. We use a sequence-to-sequence model to predict future paths from observed paths and we incorporate class information into the model by concatenating extracted label representations with traditional location inputs. We experiment with both LSTM and transformer encoders and we use generative adversarial network as introduced in Social GAN \cite{socialgan} to learn the multi-modal behavior of traffic agents. We train our model on Stanford Drone dataset which includes 6 classes of road agents and evaluate the impact of different model components on the prediction performance in multi-class scenes.

\end{abstract}

%%%%%%%%% BODY TEXT
\section{Introduction}

Predicting road agents’ trajectories is important for self-driving automobiles and robots that frequently interact with humans. This is a challenging task as road agents make their path decisions considering a variety of factors including static environment, other road agents’ motion as well as various traffic rules. In addition, there is usually no single correct future prediction as multiple paths might be acceptable at the same time. Generative Adversarial Networks (GANs) have shown potential in tackling the two problems mentioned above, namely the complex behavioral patterns and the multimodal nature of trajectory prediction \cite{socialgan,sophie,bigat}. 

However, we found in our research that most prediction models only consider a single class of road agents (usually pedestrians). In reality, traffic scenes involve various classes of road agents such as pedestrians, cyclists, buses and cars. Different road agents obey different traffic rules and may take distinct paths in the same scene (imagine that a car may never take a crossroad). In particular, to the best of our knowledge, few GAN prediction models are applied to handle multi-class trajectory prediction. Therefore, we propose to incorporate class information as additional input into prediction model using generative adversarial network to examine how the addition of class information impacts the performance of the model in multi-class traffic scenes. 

The input to our algorithm is trajectory coordinates and class labels parsed from real-scene videos. We then experiment using either LSTM or transformer as encoder to output predicted trajectories (x and y locations). We use GAN to model the multimodal behavior of road agents.

\section{Related work}
\textbf{RNNs for Trajectory Prediction}.
Recurrent neural networks, because of their autoregressive nature, have been commonly used in sequence generation tasks such as image captioning, language translation and autocompletion of sentences. The task of trajectories prediction in its nature is similar to language generation: the model is given observed status of road agents and is expected to predict future steps one after another. Therefore, RNNs or LSTMs have been widely chosen to perform trajectory prediction \cite{sociallstm, socialgan}. Recently, people have found transformers to have stronger encoding abilities and the advantage of avoiding recursion \cite{attention}. The current state-of-the-art algorithm of predicting human trajectories on ETH dataset, AgentFormer, uses transformer from end to end \cite{AgentFormer}. Therefore, we also experiment using transformer structure as encoder in addition to lstm to extract information from observed trajectories.

\textbf{Handling Multiclass Information}.
Real traffic scenes involve a diverse variety of road agents including pedestrians, bicyclists, cars, etc.  Most prediction algorithms focused on a single type of actor \cite{PECNet}. MultiXNet \cite{MultiXNet} uses a single model to handle multiple actor types by separating different sets of outputs, one for each type, and add per-type losses together as final loss. This approach ensures safe and simple operations. However, since no class information is provided during the process of encoding, the model can not learn class conditional trajectory patterns but instead the averaged trajectories of different classes. Several graph attention networks \cite{JD, Trajectron} have incorporated class information into the model by concatenating class information with other inputs and applying embedding to extract features. Nevertheless, no experiment results directly demonstrate how the addition of class information impact the prediction performance. We provide class information as input into the model using the concatenation method and carry out ablation study. 

\textbf{Multimodal Modeling}.
With the consideration that a road agent may have many plausible path options and may choose any one of them, previous work have used two common approaches to learn probability distributions instead of single values of future trajectories. The first approach is to model the distribution explicitly with conditional variational autoencoders \cite{CVAE,Y-Net} . The second approach is to model stochasticity implicitly using generative adversarial networks \cite{socialgan, sophie,bigat}.  Unlike conditional variational autoencoders, GAN does not make assumptions of the probability prior and is able to model intractable distributions. Despite GAN being a powerful tool, we have detected in our experiments that GAN has several notable defects. First, to generate multiple different results for one observed trajectory, the generator needs to receive randomly generated noises as input which may lead to random results far from the ground truth. Thus, GAN needs to run multiple sampling in order to acquire most likely future paths, which makes training much slower and less stable. Second, the common discriminator \cite{socialgan} takes one path consisting of an observed path and the corresponding predicted path as input and classifies it as fake/true. In this case, the discriminator learns to classify a complete path but we may actually want it to classify the predicted path conditioned on the observed path. Therefore, the quantitative performance improvement of applying GAN might largely be due to its larger sampling which will increase the possibility of finding a path closer to the ground truth. 

\section{Problem Statement}
Our goal is to jointly learn and predict the future trajectories of all agents involved in a scene based on their previous trajectories and classes. Specifically, we want to use the observed trajectory and class label of all agents in a scene as our input. Our model will then output predicted future trajectories for all agents. The input trajectory of an agent $i$ is defined as $X_i = (x_i^t, y_i^t)$  for time steps $t = 1, \dots, t_{obs}$ and the future trajectory (ground truth) is defined as $Y_i = (x_i^t, y_i^t)$ from time steps $t = t_{obs}+1, \dots, t_{pred}$. We denote predictions as $\hat{Y}_i$. Given class label of agent $i$, we convert the semantic class into one hot vector $c_i$ with the corresponding class entry as 1. There are six different road agents in our input to the model, namely pedestrians, bycyclists, skateboarders, cars, buses and golf carts. 

\section{Methods}

\begin{figure*}
\begin{center}
\includegraphics[width = .9\linewidth]{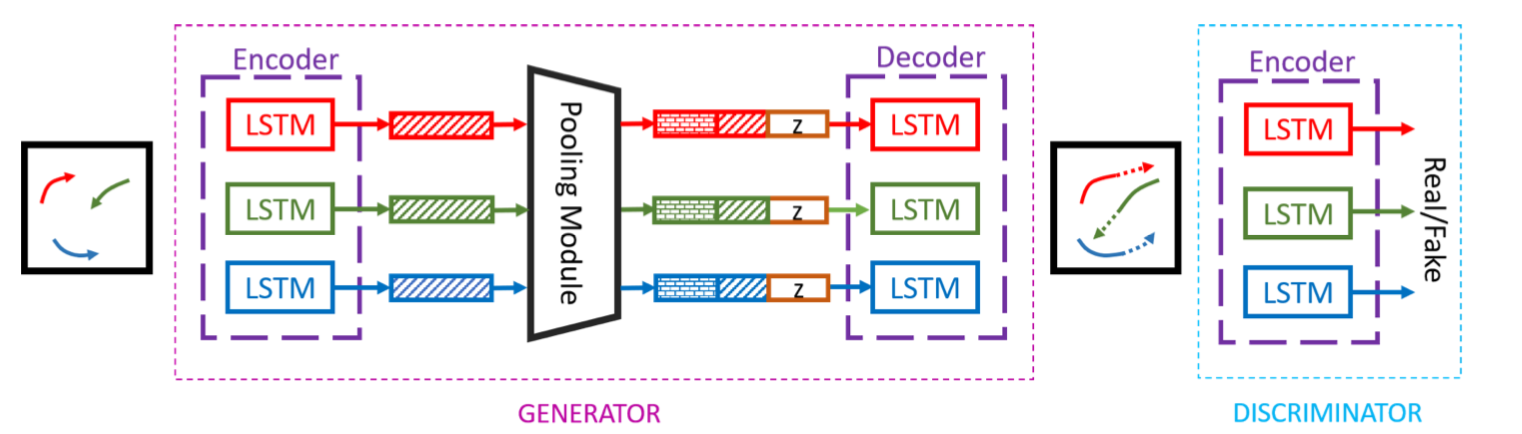}
\end{center}
   \caption{Baseline model structure}
\label{fig:short}
\end{figure*}

We use Social GAN \cite{socialgan} as the baseline and as the foundation of our approach. The model consists of three key components: Generator(G), Pooling Module(PM), and Discriminator(D) (see Figure \ref{fig:short}).  G is based on an encoder-decoder framework where the hidden states of encoder and decoder are linked via PM which models agent-agent interactions.  

G takes past trajectories $X_i$ and class labels $c_i$ as input and encodes the history of agent $i$ as $H_i^t$. The pooling module takes as input all $H_i^{t_{obs}}$ and outputs a pooled vector $P_i$ for each agent. The decoder generates the future trajectory conditioned on $H_i^{t_{obs}}$ and $P_i$. D consists of a separate encoder. It takes as input $T_{real} = [X_i, Y_i]$ or $T_{fake} = [X_i, \hat{Y_i}]$ and classifies them as real or fake. We propose to add class information into the model in the following way.

\textbf{Generator}. In the encoding stage of the generator, we concatenate the embedded location and the embedded class vector of each agent  to get a fixed length vector $e_i^t$. These fetaures are then passed into the LSTM cell at time $t$:
\begin{gather}
 s_i^t = \phi(x_i^t, y_i^t, c_i; W_{se}) \\
 c_i^t = \phi(c_i; W_{ce})\\
 e_i^t = (s_i^t, c_i^t)\\
h_{ei}^t = LSTM(h_{ei}^{t-1}, e_i^t; W_{encoder})
\end{gather}
where $\phi(\cdot)$ is a linear embedding function with $W_{se}$ as spatial embedding weight and $W_{ce}$ as class embedding weight. We plan to share the LSTM weights $W_{encoder}$ among all class types instead of using one set of weights for each semantic class. We also support replacing the lstm encoder with transformer by first embedding the feature vectors $e_i^t$ into the given encoder hidden dimension and then put it into transformer encoder to get final hidden layer. After experiments, we chose 4 multi-attention heads and 4 layers for the transformer architecture. 

Since the final hidden layer $h_i^{t_{obs}}$ already contains class information, we do not make further modification to the pooling module \cite{socialgan}. The decoding stage uses similar LSTM structure to output predicted paths:
\begin{gather}
 e_i^t = \phi(x_i^{t-1}, y_i^{t-1}; W_{ed}) \\
h_{di}^t = LSTM(h_{di}^{t-1}, e_i^t; W_{decoder}) \\
(\hat{x}_i^t, \hat{y}_i^t) = \gamma(h_{di}^t)
\end{gather}
where $\gamma(\cdot)$ is a MLP.

\textbf{Discriminator}. The discriminator takes $T_{real} = [X_i, Y_i]$, $T_{fake} = [X_i, \hat{Y_i}]$, and class labels $Ci$ as input. It passes the input into an encoder with the same structure as the one in generator which also supports transformer architecture. After the sequence which consists of both observed part and predicted part is encoded, the final hidden layer is put into a MLP to calculate its fake/real classification scores. 

Note that the proposed method of adding class information is equivalent to conditional GAN which provides both the generator and discriminator with additional input c \cite{conditionalGAN}.

\textbf{Loss}. The discriminator loss is calculated using the binary cross entropy loss measuring how well the discriminator tells the difference between fake and real sequences:
\begin{gather}
\ell_D = -\mathbb{E}_{x \sim p_\text{data}}\left[\log D(x)\right] - \mathbb{E}_{z \sim p(z)}\left[\log \left(1-D(G(z))\right)\right]
\end{gather}

The generator adversarial loss measures how well the generated paths fooled the discriminator:
\begin{gather}
\ell_G  =  -\mathbb{E}_{z \sim p(z)}\left[\log D(G(z))\right]
\end{gather}

In addition to adversarial loss, we also apply L2 loss of predicted coordinates of the trajectories to the generator. To encourage the model to produce diverse samples, for each observed trajectory, we generate  k predicted paths by injecting random noise into the decoder and the L2 loss is calculated on the prediction closest to ground truth. 
\begin{gather}
L_{variety}  =  min_k \parallel Y_i - \hat{Y}_i^{(k)} \parallel_2
\end{gather}
Here we use k=20 as suggested in SGAN.

Using this loss, the generator will be able to learn from both the discriminator and the true trajectories, which is desirable. We train discriminator and generator iteratively during training. 

\section{Data and Features}
The baseline model Social GAN originally used two datasets ECH and UCY. Both datasets involve only pedestrians and scenarios are captured at 2.5Hz. To evaluate the performance of our new model, we use the Stanford Drone Dataset \cite{drone}. The Stanford Drone Dataset collects $19K$ images of various types of agents (not just $11.2K$ pedestrians, but also $6.4K$ bicyclists,  $0.3K$ skateboarders,  $1.3K$ cars,  $0.1K$ buses, and  $0.2K$ golf carts) and targets are annotated with their class label and trajectory. The Stanford Drone dataset is large ($69$G), so we first parse it to only obtain timestamp, class labels, and $x$, $y$ coordinates from the bounding box. Then, we split the data into training, testing, and validation parts using a ratio of $8:1:1$. 
\begin{figure}[ht]
    \centering
    \includegraphics[width=6cm]{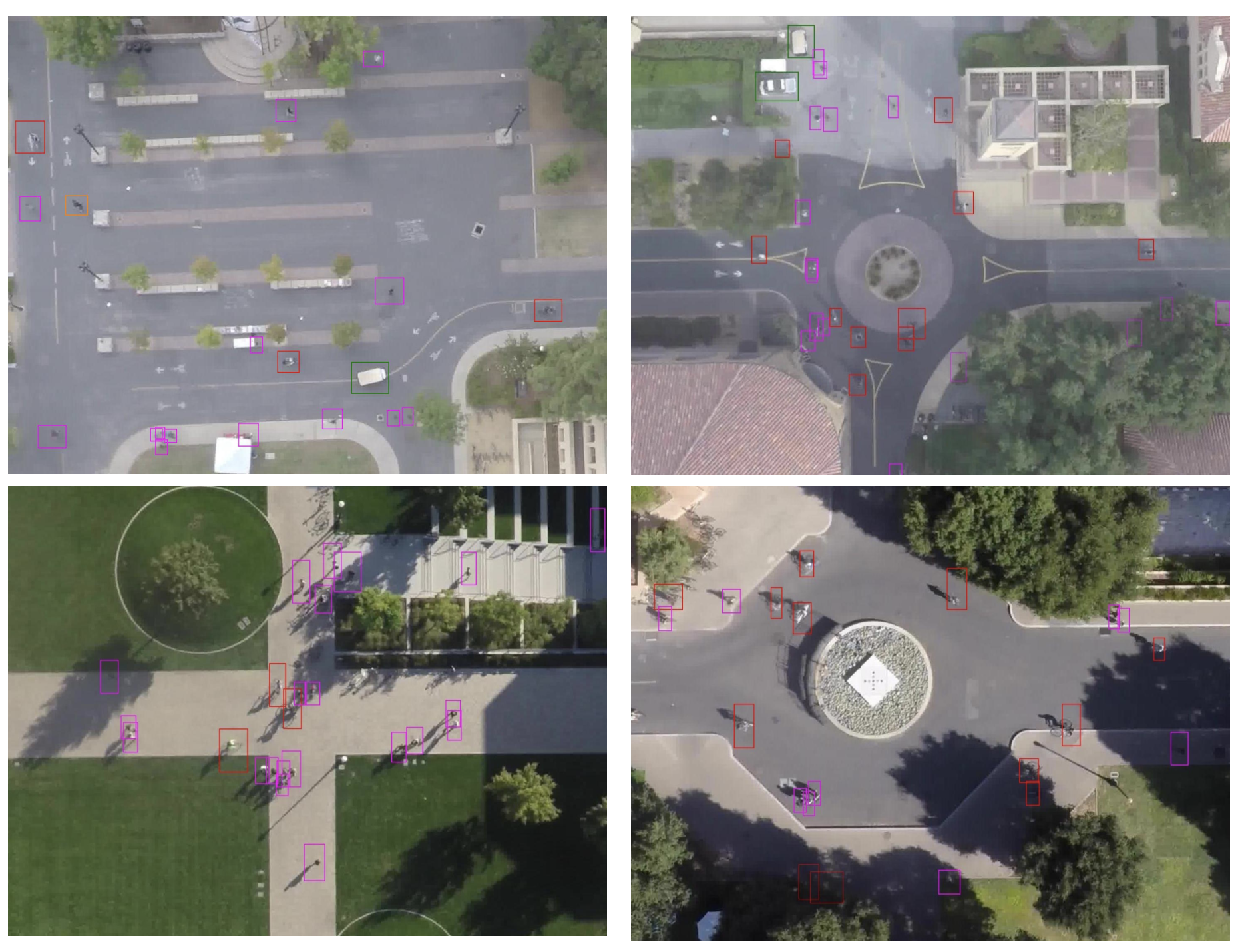}
    \caption{Sample images from Stanford Drone Dataset} \cite{drone}
    \label{fig:my_label}
\end{figure}

\begin{figure}[H]
    \centering
    \includegraphics[width=9cm]{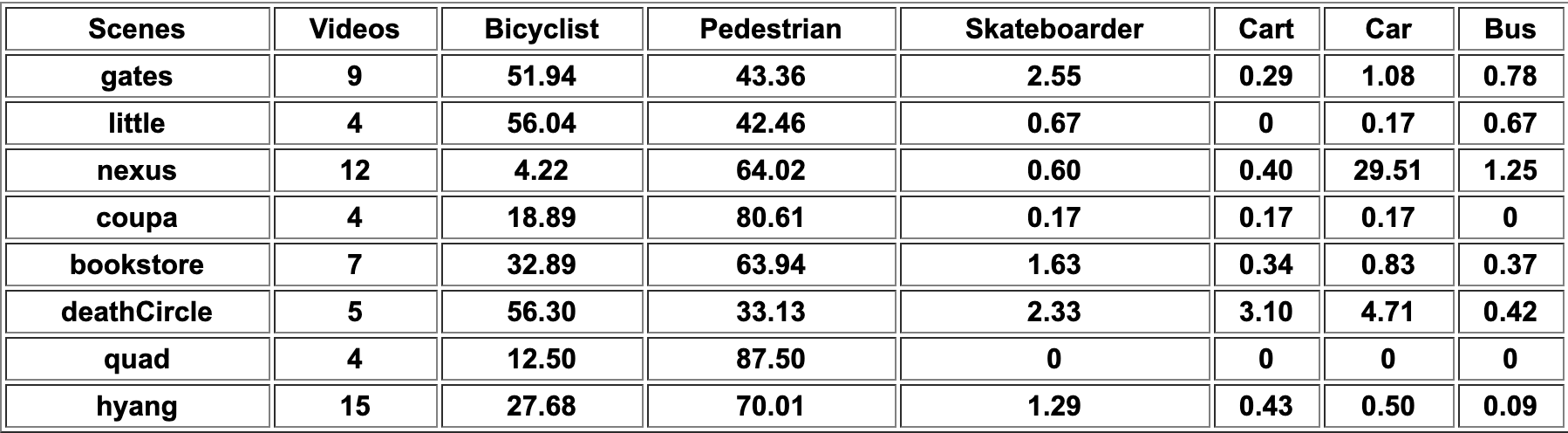}
    \caption{Breakdown of Percentages of Different Classes in Different Scenes } \cite{drone}
    \label{fig:dataset}
\end{figure}

\section{Experiments/Results/Discussion}
As mentioned in the previous section, we train our model using the Stanford Drone dataset which contains label information. To account for the possible effects of this new label feature, we add a label embedding layer in our encoder. We also experiment with another change in the encoder: we replace the LSTM layer with the transformer to see if the use of multi-head attention could help improve our model performance. To evaluate whether our above two changes are effective, we compare the training and testing results among the following four conditions: model with/without labels, and model with/without the transformer. Throughout the experiment, we use hyperparameters similar to which in the SGAN paper\cite{socialgan}. We choose a batch size of 48, learning rate of 1e-3 for both the generator and discriminator with Adam optimizer. All models are trained on the drone dataset for 200 epochs.

We use two metrics for final evaluation of the model: 
\begin{itemize}
    \item  Average Displacement Error (ADE): the average of the root mean squared error (RMSE) between the ground truth and model predictions among all trajectories $T$ in the dataset $D$.
    \begin{gather}
        ADE_D = |D|^{-1}\sum_{T\in D}RMSE_T\\
        RMSE_T=\sqrt{|T|^{-1}\sum_{(x_i, y_i)\in T}(x_i-\hat{x}_i)^2+(y_i-\hat{y}_i)^2}
    \end{gather}
    where $\hat{x}_i, \hat{y}_i$ denote the predictions of the coordinates made by the model, $|D|$ denote the number of trajectories in the dataset and $|T|$ denote the number of points in trajectory $T$. A low ADE shows the predicted trajectory made by the model is on average close to the true trajectory .
    \item Final Displacement Error (FDE): the root mean squared error (RMSE) between the ground truth and the model predictions of the final displacement among all trajectories $T$.
    \begin{gather}
        FDE_D=\sqrt{|D|^{-1}\sum_{T\in D}(T^{-1}_x-\hat{T}^{-1}_x)^2+(T^{-1}_y-\hat{T}^{-1}_y)^2}
    \end{gather}
    where $(T^{-1}_x, T^{-1}_y)$ denote the coordinate of the last point in trajectory $T$ and $(\hat{T}^{-1}_x, \hat{T}^{-1}_y)$ be the model's prediction.
    
\end{itemize}

During the training process of the original social GAN model without label embedding, we discover that the discriminator is not well-trained. The adversarial of the generator is nearly constant and the discriminator data loss is consistently high. 
\begin{figure}[H]
    \centering
    \includegraphics[width=6cm]{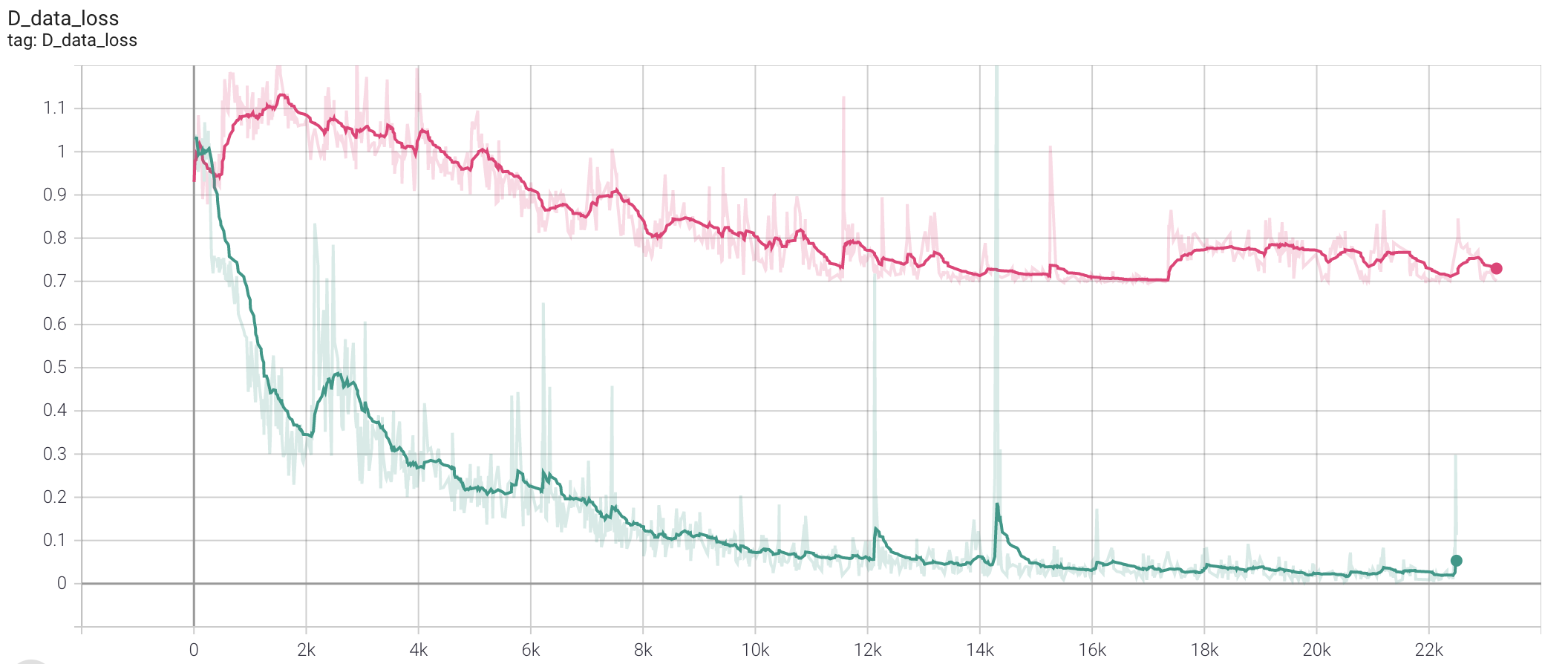}
    \caption{Discriminator data loss for original Social GAN model with ReLU activation (red) and leakyReLU(green)}
    \label{fig:Dloss}
\end{figure}
\begin{figure}[H]
    \centering
    \includegraphics[width=6cm]{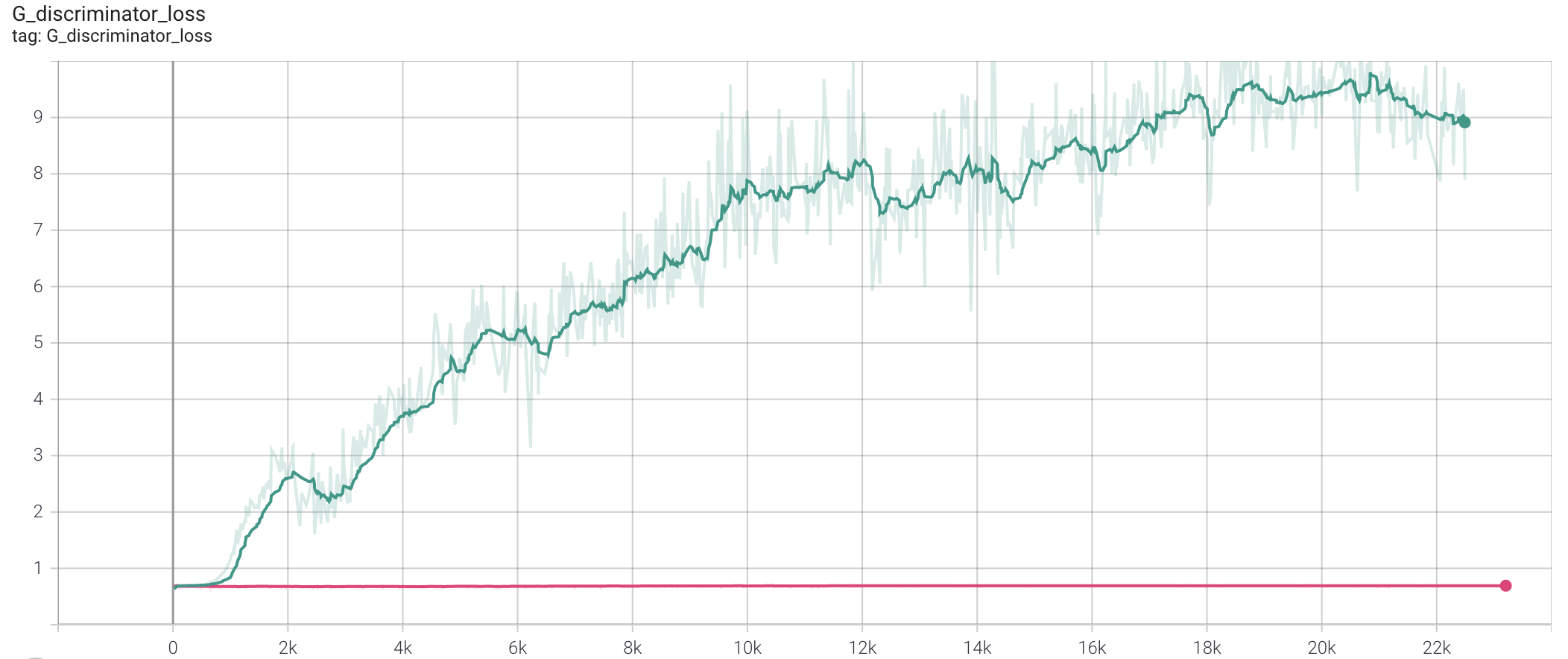}
    \caption{Generator adversarial loss for original Social GAN model with ReLU activation (red) and leakyReLU(green)}
    \label{fig:Gloss}
\end{figure}
This is because the encoder of the discriminator of the original SGAN model uses ReLU activation. As a result, nodes of the discriminator become inactive during training. We thus decide to use leakyReLU activation in place of ReLU to fix the vanishing gradient issue. See Figure \ref{fig:Dloss} , Figure \ref{fig:Gloss} for the plots of the Discriminator data loss and Generator adversarial loss with ReLU activation vs leakyReLU activation. The model with leakyReLU activation outperforms the model with ReLU activation on both metrics, with the leakyReLU model having ADE/FDE equal to 23.05/45.73 compared to the ReLU model having ADE/FDE equal to 23.56/46.86. However, one thing to notice is that the adversarial loss of the generator is, in fact, increasing. Observing this increment of loss, we suspect that the generator is trained mostly based on the L2 loss. Therefore, we also experiment with models with GAN removed from the architecture. In other words, we only use the generator and its L2 loss to train our model. It turns out that our model without GAN performs almost equally well as the other models with GAN. However, the training time is much faster than GAN models. The original SGAN model needs 7 hours and 30 minutes of training with the parameters specified above to converge on a Tesla K80 GPU. However, using the same parameters, the model without GAN requires only 1 hour and 30 minutes of training to converge in the same environment. \\

The following chart shows the results of our experiments. We use an observation length of 8 frames (3.2 seconds) and prediction length of 12 frames (4.8 seconds) to evaluate our models. Unless otherwise indicated, the model uses leakyReLU activation:

\begin{center}
 \begin{tabular}{||c c c||} 
 \hline
 Model & ADE & FDE \\ 
 \hline\hline
  original SGAN (ReLU activation)  & 23.56 & 46.86 \\ 
 \hline
 GAN (LeakyReLU activation)  & \textbf{21.98} & \textbf{43.53} \\
 \hline
 GAN with label   & 23.05 & 45.73 \\
 \hline
 GAN with transformer  & 23.06 & 45.77 \\ 
 \hline
 noGAN  & 23.02 & 46.83 \\
 \hline
 noGAN with label& 23.00 & 47.19 \\
  \hline
 noGAN with transformer  & 22.73 &  46.90 \\
%  \hline
%  nolable\_notransformer & 12 & 23.59 & 46.88\\
 \hline
\end{tabular}
\end{center}
While all models have similar performance scores, the model without label embedding has the best performance, with an ADE/FDE equal to 21.98/43.53. We think that this is because LSTM and transformer encoder layers are strong enough to infer label information from the observed trajectories. In addition, since scenes on the dataset are diverse, the road agents with the same label behave differently on different scenes. So the information of the label may not be so useful. The label embedding increases the hidden dimension size, which makes the model harder to train. Thus the models with the label information will not necessarily be better.\\

\begin{figure}[H]
    \centering
    \includegraphics[width=6cm]{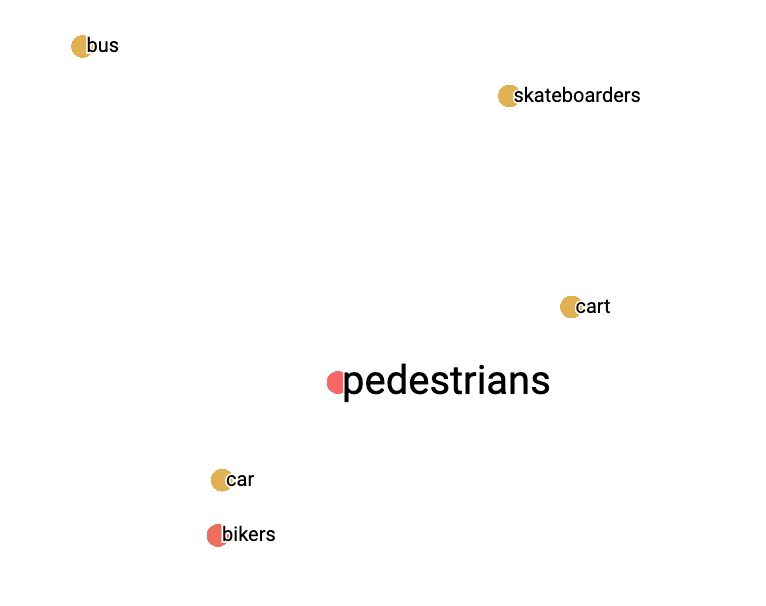}
    \caption{Projection of label embedding using PCA }
    \label{fig:pca}
\end{figure}

\begin{figure}[H]
    \centering
    \includegraphics[width=6cm]{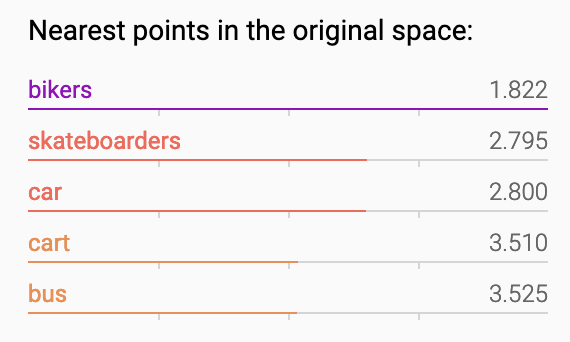}
    \caption{Euclidean distances of pedestrians from other classes}
    \label{fig:dist}
\end{figure}

We plot the projection of the label embedding vectors using PCA(Principal Component Analysis) in Figure \ref{fig:pca}. Figure \ref{fig:dist} shows the Euclidean distances of the embedding for pedestrians from other classes. Figure \ref{fig:pca} shows that embedding of cars and bikers are close, as they both occur on roads and their trajectory are similar. From Figure \ref{fig:dist}, we can see that pedestrian embedding is close to bikers and skateboarders and far from cart and bus, which is desired. 

\begin{figure}[H]
    \centering
    \includegraphics[width=6cm]{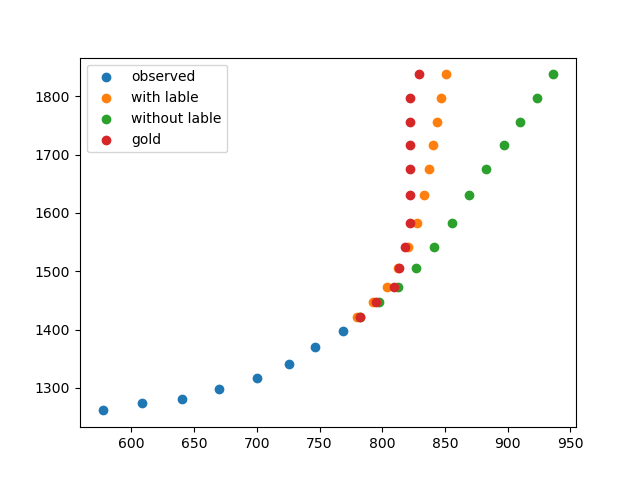}
    \caption{trajectories predictions of model with label information vs without label information}
    \label{fig:compareLable}
\end{figure}

We now begin to analyze some prediction examples of the models. First, comparing the predictions of the model with transformer encoder and LSTM encoder. We find that although they have similar ADE/FDE scores, the transformer encoder makes more reasonable predictions in many situations. We plot 30 trajectory examples where either model had large L2 error. While the model with transformer encoder gives reasonable predictions throughout the examples, we find that the model with LSTM encoder exhibit some less reasonable predictions. Figure \ref{fig:compare} shows the trajectory prediction of the model with transformer encoder and LSTM encoder on the same observed trajectory:

\begin{figure}[H]
    \centering
    \includegraphics[width=6cm]{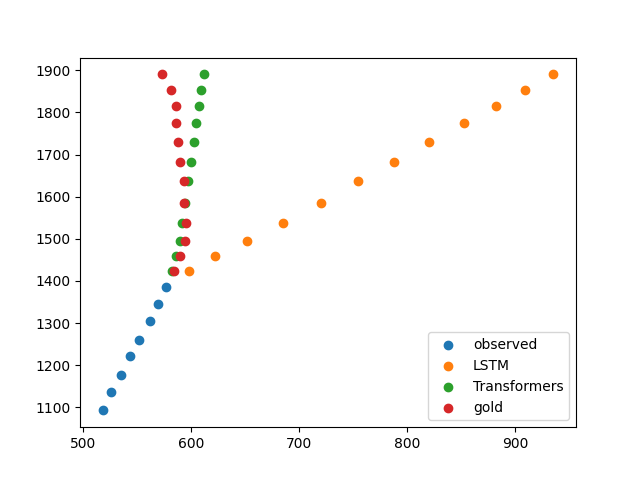}
    \caption{predicted trajectories of model with transformer encoder vs LSTM encoder}
    \label{fig:compare}
\end{figure}
On the one hand, we observe that there is a sharp turn in the trajectory predicted by the LSTM encoder, which is not very reasonable. On the other hand, the trajectory predicted by the transformer is smoother and closer to the true trajectory. \\
Similarly, we compare the predictions of the model with and without label embeddings. We see that although model without label have a better ADE/FDE score, the model with label makes more reasonable predictions in certain situations. Figure \ref{fig:compareLable} shows the trajectory of a biker in the scene "deathcircle":\\
The scene "deathcircle" contains a crossing with roundabout for bikes. While the pedestrians in the scene have more variable trajectories, the bikers in the scene mostly have trajectories corresponding to "left turn", "right turn" and "crossing" at the roundabout. So the model with label information correctly predicts that the observed biker in Figure \ref{fig:compareLable} is making a left turn at the roundabout. However, the model without label information predicts that the road agent will continue going forward smoothly, which is impossible for a biker in this scene.\\

\begin{figure}[H]
    \centering
    \includegraphics[width=6cm]{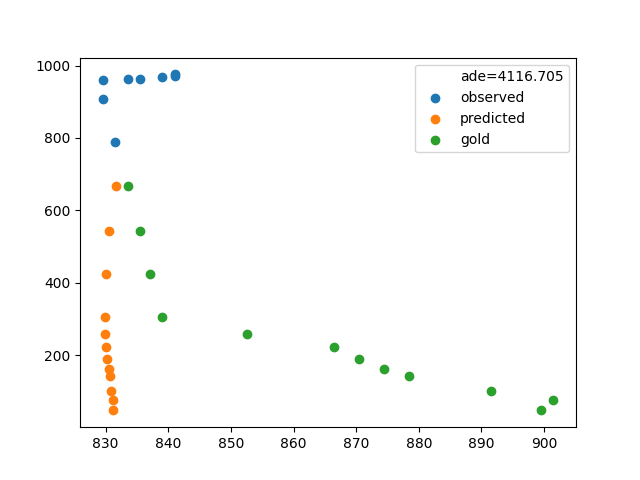}
    \caption{worst predicted trajectories of model with LSTM encoder}
    \label{fig:LSTMworst}
\end{figure}

\begin{figure}[H]
    \centering
    \includegraphics[width=6cm]{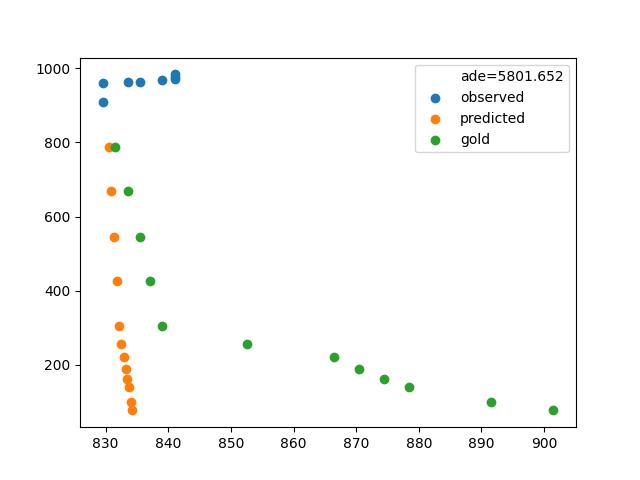}
    \caption{worst predicted trajectories of model with transformer encoder}
    \label{fig:Transworst}
\end{figure}

To investigate the reasoning behind the large errors, we pick 30 worst predicted trajectories for the model with transformer encoder and LSTM encoder, respectively. The prediction of both of the models are reasonable in all examples. Interestingly, we found a common example for both models with worst prediction trajectories. Figure \ref{fig:LSTMworst} shows the predicted trajectory of model with LSTM encoder and Figure \ref{fig:Transworst} shows the predicted trajectories for the model with transformer encoder. \\
While both models achieved very large ADE loss in this example, they made similar prediction trajectories: predicting that the road agent will continue forward after the sharp turn. However, the true trajectory of the road agent deviates from going straight after the sharp turn and makes another sharp left turn afterwards. It is hard even for human to predict the trajectories in this example. Therefore, the predictions of both models are reasonable.

\section{Conclusion/Future work}
The model using GAN training, LSTM encoder, leakyReLU activation, and no label embeddings has the best performance among all models. Although the addition of class labels and transformers into the encoder does not improve the model's performance quantitatively, we discovered and analyzed certain situations when those designs enable the model to make better predictions. 

One reason of class labels not contributing to the improvement of quantitative results might be due to the nature of the dataset. The majority of road agents in Stanford Drone dataset are pedestrians and bikers as shown in \ref{fig:dataset} and the existence of small proportion of carts and buses might actually be treated as noise and disturbs the prediction. To tackle this problem, one might consider make better and more averaged selection of the dataset. Another possible improvement to make is to have separate decoders for each class instead of just injecting class label as additional input.

Since our models uses only the trajectory coordinates to make predictions, the model does not have information about the surrounding environment of road agents. In the future, we want to add scene information to the model. In particular, we can add image embedding of the scene image to the input.

\section{Contributions $\&$ Acknowledgements}
We used published code of SGAN \cite{socialgan} as our baseline. The github link is https://github.com/agrimgupta92/sgan.git. J.L. did preliminary research and literature review. T.C. parsed dataset. A.L., J.L. supervised training and designed model improvements. A.L. generated plots and analyzed qualitative results. J.L., A.L., T.C. wrote the paper. T.C. made the slide for presentation.

{\small
\bibliographystyle{ieee_fullname}
\bibliography{main}
}

\end{document}